\documentclass[final]{cvpr}

\usepackage{times}
\usepackage{epsfig}
\usepackage{graphicx}
\usepackage{amsmath}
\usepackage{amssymb}
\usepackage{array}
\usepackage{subcaption}
\usepackage{tabularx}
\usepackage{adjustbox}
\usepackage{multirow}
\usepackage[dvipsnames]{xcolor}
\usepackage{float}
\usepackage{multicol}
\newcommand{\Sb}{\mathbf{S}}
\newcommand{\Shb}{\mathbf{\hat{S}}}
\newcommand{\rkh}{\mathbf{\hat{R}}}

\usepackage[pagebackref=true,breaklinks=true,colorlinks,bookmarks=false]{hyperref}

\makeatletter
\newcommand{\printfnsymbol}[1]{%
  \textsuperscript{\@fnsymbol{#1}}%
}

\def\cvprPaperID{80} 
\def\confYear{CVPR 2021}
\begin{document}

\title{(ASNA) An Attention-based Siamese-Difference Neural Network with Surrogate Ranking Loss function for Perceptual Image Quality Assessment}

\author{Seyed Mehdi Ayyoubzadeh \thanks{The authors have contributed equally}\\
McMaster University\\
Hamilton, Canada\\
{\tt\small ayyoubzs@mcmaster.ca}
\and
Ali Royat \textsuperscript{*}\\
Sharif University of Technology\\
Tehran, Iran\\
{\tt\small royat.ali@ee.sharif.edu}
}

\maketitle

\begin{abstract}
Recently, deep convolutional neural networks (DCNN) that leverage the adversarial training framework for image restoration and enhancement have significantly improved the processed images' sharpness. Surprisingly, although these DCNNs produced crispier images than other methods visually, they may get a lower quality score when popular measures are employed for evaluating them. Therefore it is necessary to develop a quantitative metric to reflect their performances, which is well-aligned with the perceived quality of an image. Famous quantitative metrics such as Peak signal-to-noise ratio (PSNR), The structural similarity index measure (SSIM), and Perceptual Index (PI) are not well-correlated with the mean opinion score (MOS) for an image, especially for the neural networks trained with adversarial loss functions. 
   This paper has proposed a convolutional neural network using an extension architecture of the traditional Siamese network so-called Siamese-Difference neural network. We have equipped this architecture with the spatial and channel-wise attention mechanism to increase our method's performance.
    Finally, we employed an auxiliary loss function to train our model. The suggested additional cost function surrogates ranking loss to increase Spearman's rank correlation coefficient while it is differentiable concerning the neural network parameters.
Our method achieved superior performance in \textbf{\textit{NTIRE 2021 Perceptual Image Quality Assessment}} Challenge. The implementations of our proposed method are publicly available.\footnote{\href{https://github.com/smehdia/NTIRE2021-IQA-MACS}{https://github.com/smehdia/NTIRE2021-IQA-MACS}}
\footnote{\href{https://github.com/AliRoyat/MACS-IQA-Pytorch}{https://github.com/AliRoyat/NTIRE2021-IQA-MACS-Pytorch}}

\end{abstract}

\section{Introduction}

\begin{figure*}[h!]
  \begin{subfigure}[c]{0.21\linewidth}
    \includegraphics[width=\linewidth]{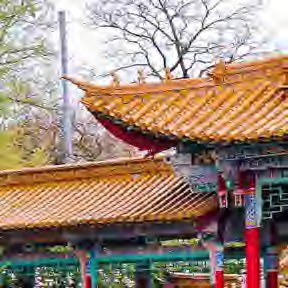}
    \caption{image A}
  \end{subfigure}
    \begin{subfigure}[c]{0.21\linewidth}
    \includegraphics[width=\linewidth]{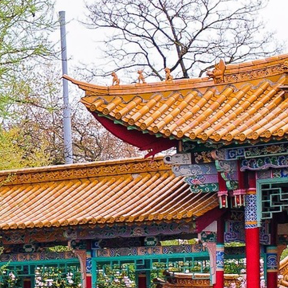}
    \caption{reference image}
  \end{subfigure}
    \begin{subfigure}[c]{0.21\linewidth}
    \includegraphics[width=\linewidth]{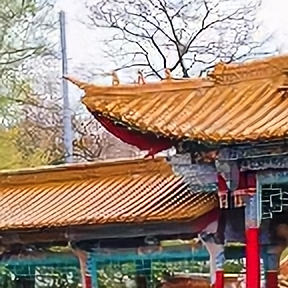}
    \caption{image B}
  \end{subfigure}
  \begin{subfigure}[c]{0.16\linewidth}
    \begin{tabular}{ c|c|c } 
 Method & A & B \\
 \hline
 PNSR & \textcolor{Blue}{$21.13$} & \textcolor{Red}{$19.51$} \\ 
 SSIM & \textcolor{Blue}{$0.77$} & \textcolor{Red}{$0.77$} \\ 
 MS-SSIM & \textcolor{Blue}{$0.93$} & \textcolor{Red}{$0.91$} \\
 NIQE & \textcolor{Blue}{$28.07$} & \textcolor{Red}{$23.70$} \\
  UQI & \textcolor{Blue}{$0.97$} & \textcolor{Red}{$0.96$} \\
 VIFP & \textcolor{Blue}{$0.24$} & \textcolor{Red}{$0.22$} \\
 ASNA (Ours) & \textcolor{Red}{$1464.08$} & \textcolor{Blue}{$1473.48$} \\
 MOS & \textcolor{Red}{$1396.14$} & \textcolor{Blue}{$1533.14$} \\
\end{tabular}
  \end{subfigure}
  \caption{As illustrated, image B is similar to the reference image than image A visually (in fact, its MOS score is higher); however, all of the methods except ours prefer image A over B}
  \label{fig:fig0}
\end{figure*}

DCNNs have shown their effectiveness in a wide range of computer vision and image processing tasks, including single-image super-resolution, denoising, deblurring, etc. \cite{Albluwi2018, 1707.02921, 1501.00092}. The major drawback of traditional DCNNs is that they often produce over smooth images for rich textured images, mainly due to the improper metrics for training DCNN. \\ Emerging of the Generative Adversarial Networks (GAN) \cite{1406.2661} did an evolutionary step for learning distributions with sharp peaks. 
Researchers adopted the adversarial training framework for image restoration tasks. Using adversarial loss facilitates the DCNN ability to produce sharp and crispy images. The produced images by GAN-based DCNNs are typically more pleasant to human eyes than the DCNNs that do not use adversarial loss.  However, such networks often get a lower score than the plain DCNNs when famous metrics are used for comparison \cite{1609.04802} while their MOS are higher than their counterparts. 
The reason is that most of the DCNNs use Mean Squared Error (MSE) as the objective function. Therefore, they get a higher peak signal-to-noise ratio (PSNR) \cite{Hore2010} as PSNR is directly related to the MSE. 
PSNR and the structural similarity index measure (SSIM) \cite{Wang2004} are the most common metrics for full reference image quality assessment (IQA).  Such metrics' efficacy is disappointing, especially for evaluating fine textures and details in the images \cite{1609.04802}. Note that the ultimate goal of image enhancement networks is to generate visually pleasurable images for humans and have a high MOS.
So developing a new full reference metric for IQA is necessary for comparing the different proposed methods for image enhancement tasks and optimizing DCNNs with a measure that is highly correlated with the perceived quality assessment of humans.  \\
The DCNNs can extract information about underlying structures and features in the images, and thus, they can be a powerful metric for full reference IQA provided there is enough data to train them. Fortunately, public datasets such as TID 2008 \cite{Ponomarenko2004TID2008A},  TID 2013 \cite{PONOMARENKO201557}, PieApp \cite{prashnani2018pieapp} and PIPAL \cite{pipal} provide images with their corresponding reference images and MOS to train DCNNs in a supervised manner. \\
There are two desired characteristics for the full reference IQA metrics:
(i) high Pearson linear correlation coefficient  (PLCC) \cite{2008} between the scores produced by the proposed method and MOS, which indicates the linear relationship between them, (ii) high Spearman's rank correlation coefficient (SRCC) \cite{1}, which shows the monotonicity of relationship between the proposed method and MOS.
In this paper, we have proposed an attention-based Siamese-Difference neural network architecture for IQA, dubbed (ASNA). Siamese-Difference neural network architecture is robust for extracting the difference between two images. It is an asymmetric extension of the Siamese neural network \cite{Koch2015SiameseNN} that was initially used for scene change detection \cite{CayeDaudt2018}.
We have equipped our Siamese-Difference convolutional neural network with an attention mechanism \cite{1705.02544} to make the network able to highlight the differences between an input image and its corresponding reference image.
Besides, we have integrated PLCC directly into the loss function to increase the PLCC of the proposed method. Sadly, we could not do the same with SRCC since it is a non-differentiable function concerning the neural network parameters because of its ranking operation. To circumvent this issue, we have used the method proposed in SoDeep \cite{1904.04272} and trained a neural network that can surrogate ordering operation while it is differentiable. Then, we have added the difference between the outputs for MOS and the scores given by ASNA. This auxiliary loss function approximates the ranking difference between MOS and the outputs of ASNA, which is directly correlated with SRCC.
 The experimental results show our method outperformed other metrics for full reference IQA, and it has a significantly greater PLCC and SRCC compared to other measures.
Our  methods  ranked  \textit{\textbf{9th}}, in \textbf{\textit{NTIRE 2021 Perceptual Image Quality Assessment}} \cite{gu2021ntire}. 
Our key contributions are as follows:
\begin{itemize}
    \item Proposing a Siamese-Difference Neural Network architecture equipped with attention mechanism that is powerful for focusing on the difference between an input image and the reference image.
\item Using an auxiliary differentiable surrogate ranking loss function to improve SRCC.
\item  Extensive experiments manifest the superiority of ASNA for full reference IQA.
\end{itemize}

\begin{figure}[H]
  \centering
    \begin{subfigure}[c]{0.4\linewidth}
    \includegraphics[width=\linewidth]{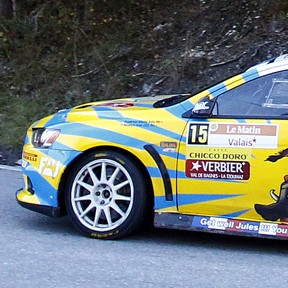}
    \caption{}
  \end{subfigure}
    \begin{subfigure}[c]{0.4\linewidth}
    \includegraphics[width=\linewidth]{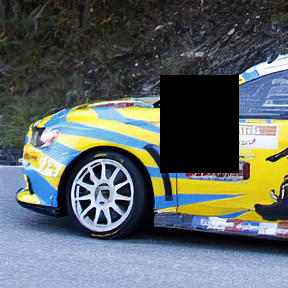}
    \caption{}
  \end{subfigure}
      \begin{subfigure}[c]{0.4\linewidth}
    \includegraphics[width=\linewidth]{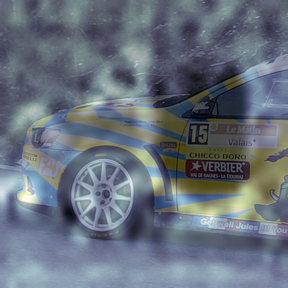}
    \caption{}
  \end{subfigure}
      \begin{subfigure}[c]{0.4\linewidth}
    \includegraphics[width=\linewidth]{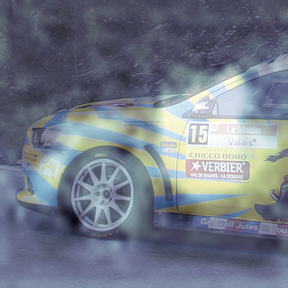}
    \caption{}
  \end{subfigure}
  \caption{(a) reference image (b) distorted image (c, d) spatial attention maps}
  \label{fig:figasnaattention}
\end{figure}

\section{Related Works}
There are numerous methods for full reference IQA. Though, most of them have a very weak correlation with the human visual system (HSV).
Standard IQA metrics hand-crafted to extract some certain statistical differences between the reference and distorted images. These designed measures are ineffective for judging GAN-based results as they do not consider degradations of GANs. 

Full reference IQA methods fall into two categories.
The methods in the first category are non-machine learning-based methods. These are the metrics handcrafted by researchers to assess image quality. The most common metric among those is PSNR. PSNR considers the pixel-wise difference between two images. This metric is far from human judgment since it may significantly change by small distortions of the reference image, including translation, rotation, and intensity scaling. In \cite{wang2002universal}, Wang et al. proposed a metric, called UQI, to decompose an image into contrast, luminance, and structure and exploit them to measure two images' similarity. The main problem of UQI is that it may be unstable under certain conditions. To mitigate this issue, SSIM \cite{Wang2004} is proposed. To improve, the correlation of SSIM with human perception from image quality, some other variants of that such as RFSIM \cite{zhang2010rfsim} , SRSIM\cite{6467149} , FSIM\cite{5705575} , VSI\cite{6873260} , MS-SSIM\cite{wang2003multiscale} and GSM \cite{6081939} are introduced. The basic idea behind all of these methods is to reformulate the decomposition formulas. 
For instance, RFSIM uses first, and second-order Riesz transforms and edge saliency feature masks to decompose an image. SRSMI uses spectral residual visual saliency (RSVS) and gradient modulus (GM). FSIM employs the combination of phase congruency and GM. 
VSI exploits visual saliency to detect the regions of an image that are more important to the human visual system. 
Likewise, in \cite{larson2010most}, Larson and Chandler propose a measure, so-called MAD, which tries to assign a score to an image based on the detection of such regions and the appearance of an image. In this method, local luminance and contrast masking are used to estimate perceived distortion in high-quality images, whereas changes in the local statistics of spatial-frequency components are employed to estimate the distortion in low-quality images.
Full reference IQA methods fall into two categories.
The methods in the first category are non-machine learning-based methods. These are the metrics handcrafted by researchers to assess image quality. The most common metric among those is PSNR. PSNR considers the pixel-wise difference between two images. This metric is far from human judgment since it may significantly change by small distortions of the reference image, including translation, rotation, and intensity scaling. In \cite{wang2002universal}, Wang et al. proposed a metric, called UQI, to decompose an image into contrast, luminance, and structure and exploit them to measure two images' similarity. The main problem of UQI is that it may be unstable under certain conditions. To mitigate this issue, SSIM \cite{Wang2004} is proposed. To improve, the correlation of SSIM with human perception from image quality, some other variants of that such as RFSIM \cite{zhang2010rfsim} , SRSIM\cite{6467149} , FSIM\cite{5705575} , VSI\cite{6873260} , MS-SSIM\cite{wang2003multiscale} and GSM \cite{6081939} are introduced. The basic idea behind all of these methods is to reformulate the decomposition formulas. 
For instance, RFSIM uses first, and second-order Riesz transforms and edge saliency feature masks to decompose an image. SRSMI uses spectral residual visual saliency (RSVS) and gradient modulus (GM). FSIM employs the combination of phase congruency and GM. 
VSI exploits visual saliency to detect the regions of an image that are more important to the human visual system. 
Likewise, in \cite{larson2010most}, Larson and Chandler propose a measure, so-called MAD, which tries to assign a score to an image based on the detection of such regions and the appearance of an image. In this method, local luminance and contrast masking are used to estimate perceived distortion in high-quality images, whereas changes in the local statistics of spatial-frequency components are employed to estimate the distortion in low-quality images. Note that the major issue with the first category's metrics is that none of them consider the context of an image properly. \\
The methods in the first category are machine learning-based methods, and they are data-driven. Since CNNs illustrated their capability in extracting underlying structures in the images, they are a suitable choice for IQA, provided that they are trained with enough data. 
LPIPS \cite{zhang2018unreasonable} , PieAPP \cite{prashnani2018pieapp} ,WaDIQaM \cite{bosse2017deep}, DISTS \cite{ding2021comparison}, SWD \cite{gu2020image} use CNNs to estimate the visual quality of an image. The first three approaches compute a distance between  two patches using weighted average of deep embedding of a CNN. 
In other words, they comprise three different parts for feature extraction, score computation, and prediction of the perceptual score.  They try to build an embedding space for the images and then compute the distance in the embedded space to measure their similarity. \\
In WaDIQaM method, the authors proposed a deep learning-based method that can be used for no-reference image quality assessment (NR-IQA). In \cite{ding2021comparison}, Ding et al. propose an specific architecture for various degradation types.  Lastly,  Gu et al. propose a metric called SWD. This metric uses the Space Warping Difference (SWD) technique to compare the features that are not only on the corresponding position but also on a small range around the corresponding position to improve the image quality's score estimation. The deep learning-based methods have a better generalization, and they are closer to human judgments in assigning quality scores to the images. However, these methods require more extension to enhance their generalization ability since they perform poorly on GAN-based results.
Note that there are also traditional techniques for NR-IQA.
There is no need for a reference image in these methods to assess an image quality score.  Among these methods can named NIQE \cite{mittal2012making}, MA\cite{ma2017learning} , PI \cite{blau2018perception} are more common than others. NIQE compares two fitted Multivariate Gaussian Model (MVG) models statistics between natural images and degraded images. The NR-IQA metrics are less useful when the reference images are available since the reference image's information can be used to examine the quality of the distorted image.

\section{Proposed Method}
\subsection{Architecture details}
Our proposed approach introduces two variants of ASNA designs, based on Siamese-Difference neural network architecture,  each trained separately, for estimating MOS. 
Siamese-Difference network design is useful for extracting minor differences between the reference image and the distorted image. \\
In the following, we describe our proposed designs:

\subsubsection{ASNA model architecture}

\begin{figure}[H]
  \centering
    \includegraphics[width=0.8\linewidth]{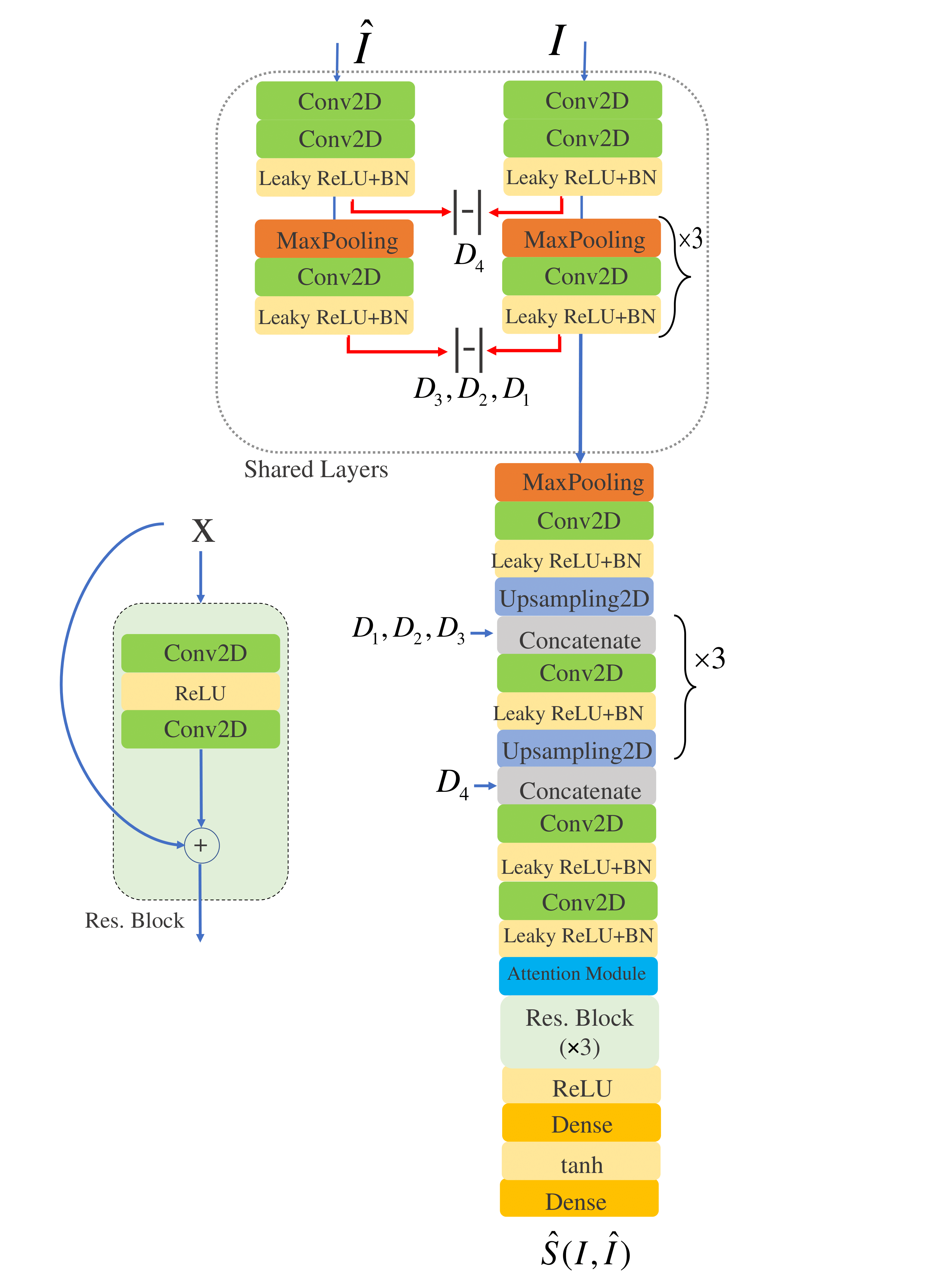}
  \caption{ASNA model architecture}
  \label{fig:figasna}
\end{figure}

\begin{figure}[H]
  \centering
    \includegraphics[width=0.8\linewidth]{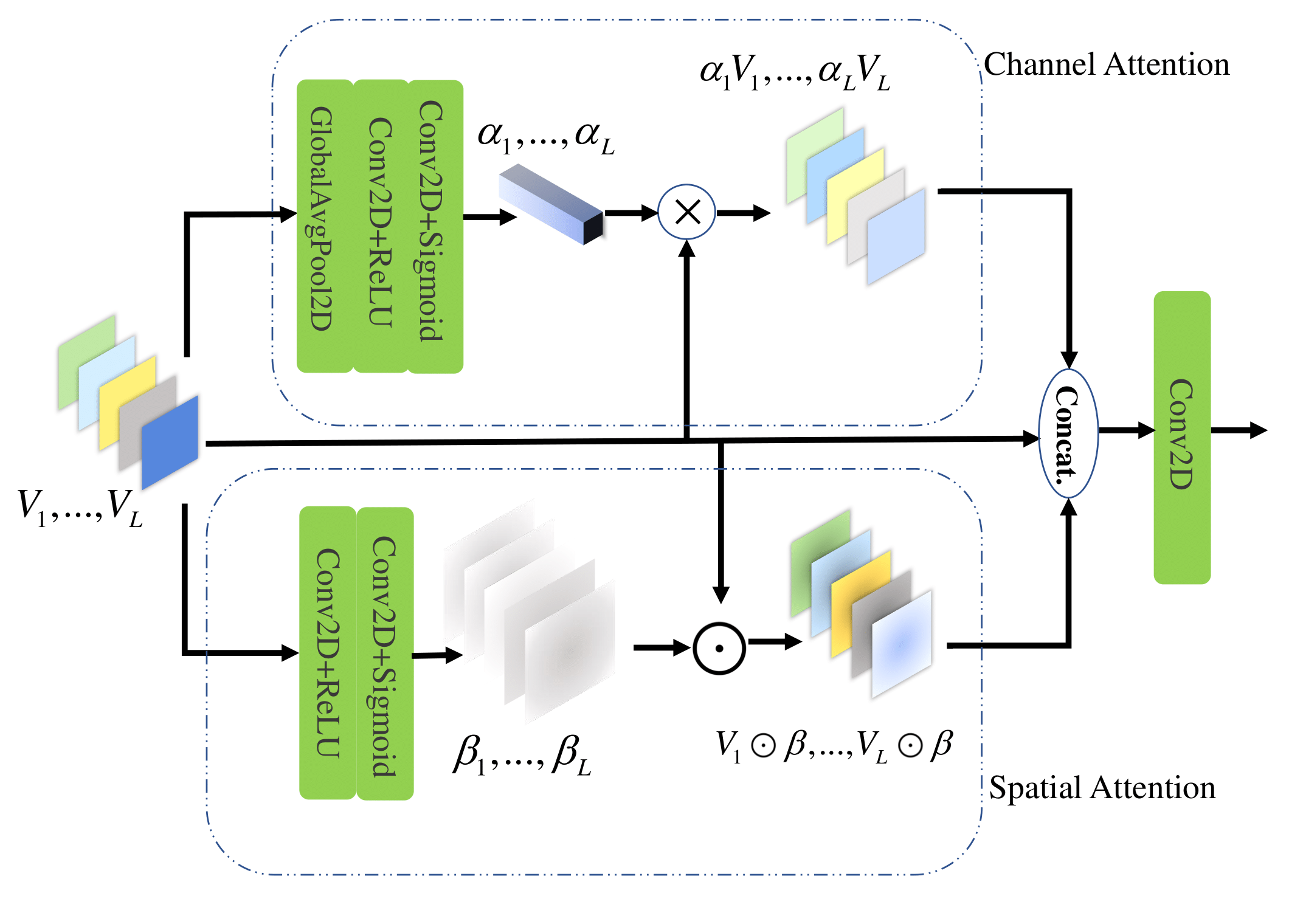}
  \caption{attention module}
  \label{fig:moduleattention}
\end{figure}

The first proposed design is shown in \ref{fig:figasna}.
As shown, the first part of the model uses shared weights for both images to encode them as in the traditional Siamese network. The encoding layers produce, $D_4$, $D_3$, $D_2$ and $D_1$ sequentially. 
In the decoding part, we concatenate the absolute difference value of the outputs from the encoding part denoted by ($D_4, D_3, D_2, D_1$).
 These difference feature maps represent the differences between the reference image and the distorted image at various feature abstraction levels. We concatenate them to the decoding part feature maps by using skip connections. 
Note that, in traditional Siamese networks, both inputs pass through a  symmetric architecture of layers. However, in ASNA, the decoding part is not symmetric for the distorted and reference images. We found out that this architecture can extract better features to estimate an image's quality score rather than the traditional Siamese network architecture.
To improve our proposed Siamese-Difference architecture capability to focus on valuable parts of the inputs, we equipped the architecture with an attention layer in the decoding part. Both channel-wise and spatial attention is used in the design of the network to improve MOS estimation. The channel-wise attention lets the network focus on the feature maps that are more important for producing the IQA score. The channel-wise attention can be seen as the soft selection of the feature maps. 
On the other hand, Spatial attention lets the networks emphasize the important spatial parts of the feature maps.
The schematic for the attention layers is visualized in \ref{fig:moduleattention}.
We have visualized some of the spatial attention maps for a sample pair of distorted and reference images in Figure \ref{fig:figasnaattention}. The attention maps are upsampled to have the same field of view as the inputs. As illustrated, the attention maps decompose the distorted parts of the image and help the network emphasize the necessary parts of the image separately.

We have also added residual blocks in this architecture to improve gradient propagation. Besides, batch normalization \cite{ioffe2015batch} is used to increase the generalization of the model. The experiments demonstrate that this architecture has significant capability to estimate the image quality score when there are enough data for learning MOS.  

\subsubsection{\small Siamese-Difference Neural Network with ConvLSTM layers}
In the previously proposed architecture, the input size was $288 \times 288$. The input size can increase the number of parameters in the fully connected layers significantly.  Also, large feature maps increase the number of computations and can be a considerable burden for the hardware. To mitigate this issue, we also propose another design to decrease the number of parameters and reduce the required computations. The proposed architecture employs ConvLSTM \cite{NIPS2015_07563a3f} layers at the input. 

\begin{figure}[H]
  \centering
  \includegraphics[width=0.7\linewidth]{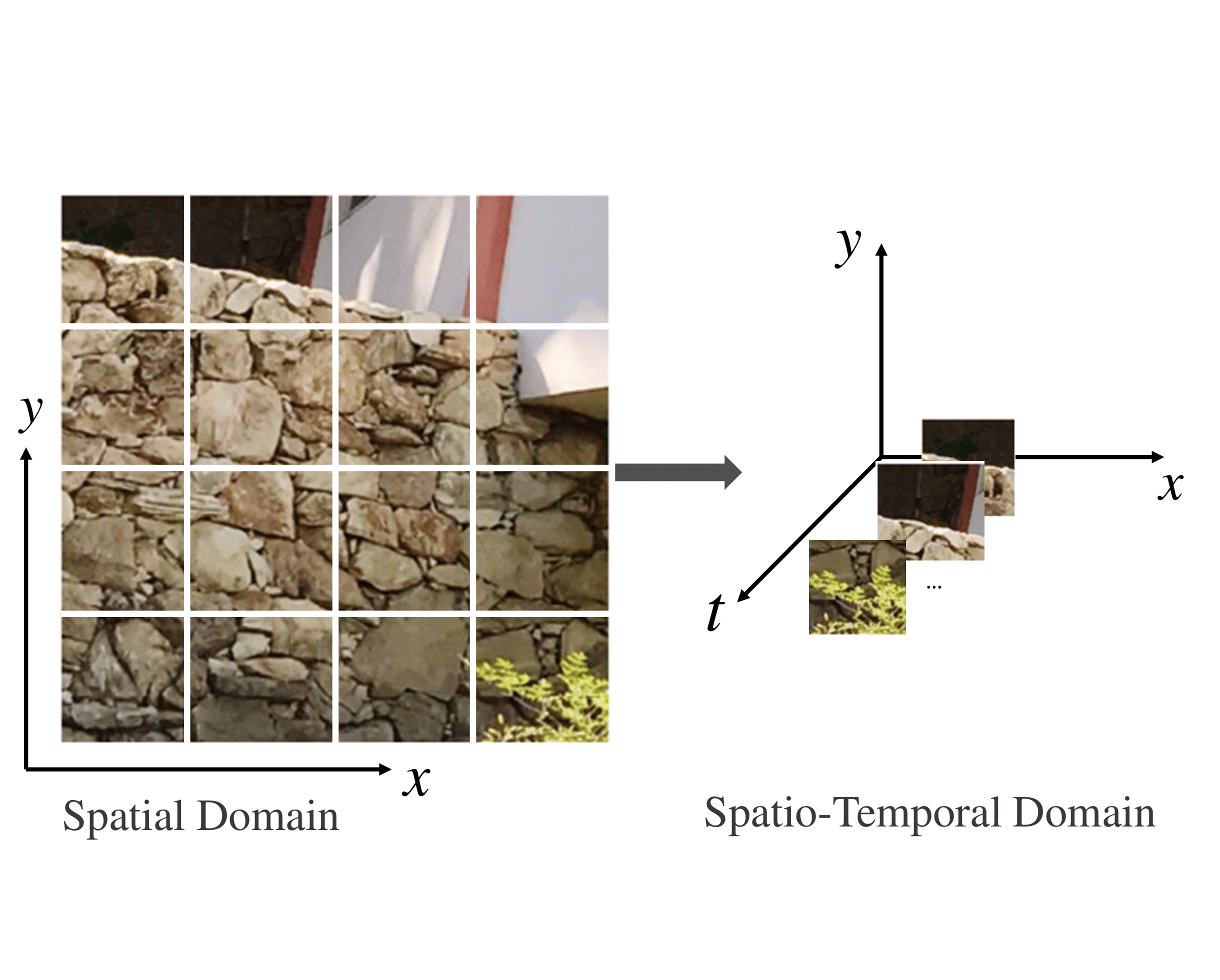}
  \caption{Tiling process}
  \label{fig:fig4}
\end{figure}

In this design, we convert the input images from the spatial to the Spatio-temporal domain by tiling the inputs. Each image is a tensor with the size of $(288, 288, 3)$; we split the image as shown in Figure \ref{fig:fig4} into $16$ tiles. These tiled images shape a new tensor with the size of $(16, 72, 72, 3)$. This tensor is passed to a ConvLSTM layer. we use the last time-step output of the ConvLSTM layer as the input feature map for the rest of the Siamese-Difference model. Note that this layer can pass the patches' useful information to the next layers of the network while limiting the architecture's number of parameters. The outputs of ConvLSTM layers are passed to the rest of the network, similar to ASNA architecture. We have also removed the attention module and residual blocks to make this model fast and efficient.
In the following section, we outline the loss function used to train the model.

\subsection{Loss function}
We have designed a specific loss function to train our architectures. Suppose the batch size is $M$, the proposed loss function for the network is:
\begin{equation}
    L=\alpha L_{MSE}+(1-\alpha) L_{Pearson}+\beta L_{Rank}
\end{equation}

\subsubsection{MSE ($L_{MSE}$)}
MSE is the standard loss function for regression tasks. MSE is the first loss function for the network to minimize the squared difference between the desired scores and the network's output.
\begin{equation}
    L_{MSE}=\frac{1}{M}\sum_{i=1}^{N}\|\Sb_i-\Shb_i\|^2
\end{equation}
Where $\Sb$ and $\Shb$ are the desired scores and the estimated scores produced by the network for $M$ inputs, respectively.

\subsubsection{Pearson's Correlation Loss ($L_{Pearson}$)}
One of the main objectives in IQA task is to increase the PLCC between the network's outputs and the ground truths. PLCC is a differentiable function concerning the neural network's parameters, so we can add that to the loss function of the network as follows:
\begin{equation}
    L_{Pearson}(\Sb, \Shb) =1-\rho_{Pearson}^2(\Sb, \Shb)
\end{equation}

Let $Cov$ and $\sigma$ denote the covariance and variance, respectively.
PLCC can be written as follows:
\begin{equation}
\rho_{Pearson}=\frac{Cov(\Sb, \Shb)}{\sigma_{\Sb}\sigma_{\Shb}}
\end{equation}

Note that this loss function can perform better than standalone $MSE$ especially when the outputs are noisy.
\subsubsection{Surrogate Ranking Loss ($L_{Rank}$)}
\begin{figure}[H]
  \centering
  \includegraphics[width=0.7\linewidth]{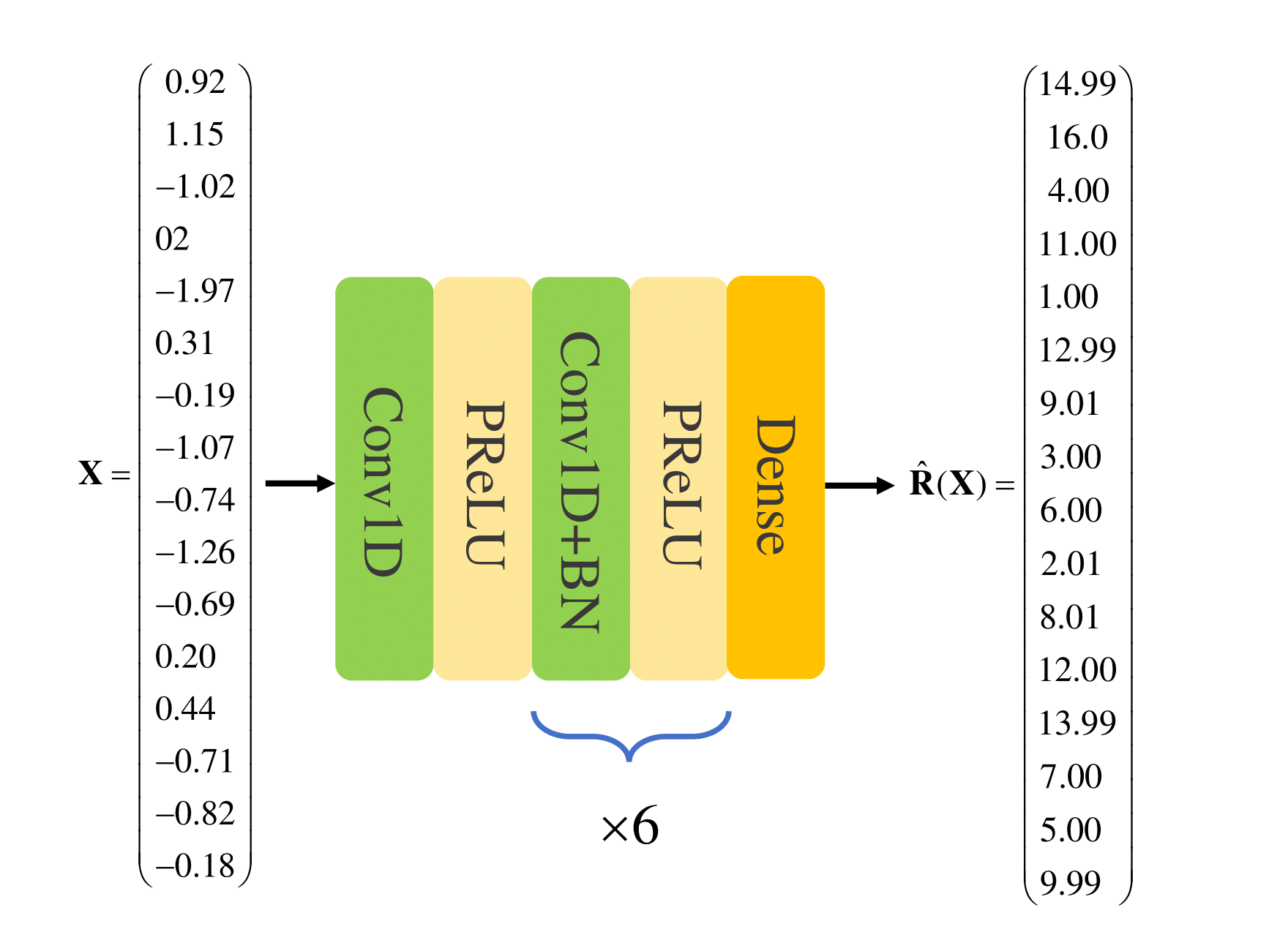}
  \caption{Ranking model architecture}
  \label{fig:fig5}
\end{figure}

One of the other performance metrics for the proposed approach is SRCC. It is desired to have a large SRCC between MOS and the network's scores. SRCC can be written as:
\begin{equation}
    \rho_{SRCC}(\Sb,\Shb)=1-\frac{6 \|\mathbf{R}(\Sb)-\mathbf{R}(\Shb)\|^2}{M(M^2-1)}
\end{equation}
Where $\mathbf{R}(\Sb)$ and $\mathbf{R}(\Shb)$  are the rank vectors of $\Sb$ and $\Shb$ respectively. Unfortunately, SRCC is a non-differentiable function since it has the ordering operation on the outputs and the ground truths. Therefore, we can not optimize this criterion directly by stochastic gradient descent. To circumvent this issue, we have used the idea in SoDeep, which is training another network to learn the rank vectors of $M$ inputs. We have employed the network architecture presented in \ref{fig:fig5}.
To train this network, we have generated $M$ dimensional random vectors and train it to learn the rank vectors corresponding to the inputs by minimize the Mean Absolute Error between the input's actual rank vector and the network's output.
An example of this network's input and output is shown in Figure \ref{fig:fig5}. In this example, the actual corresponded rank vector for the input is ($\mathbf{R}(\mathbf{X})$) $[2, 11, 9, 5, 8, 14, 4, 3, 15, 1, 7, 6, 13, 12, 16, 10]^T$. Note that the norm for the error of the surrogate ranking network ($\| \mathbf{R}(\mathbf{X})-\rkh(\mathbf{X}) \|$) is only $0.12$, which shows the surrogate network can precisely approximate the rank vector of the input while it is differentiable concerning the input. \\
 Now, we can integrate this surrogate ranking network into the training procedure of the IQA network. Suppose $\rkh(\Sb)$ and $\rkh(\Shb)$ are the estimated ranked vectors of $\Sb$ and $\Shb$, respectively. To increase the SRCC between $\Sb$ and $\Shb$, we can minimize the mean squared error between the corresponding estimated rank vectors.  This loss function can be written as follows
\begin{equation}
   L_{Rank}=\frac{1}{M}\|\rkh(\Sb) - \rkh(\Shb)\|^2
\end{equation}

\subsection{Overview of the training procedure}
\begin{figure}[H]
  \includegraphics[width=0.8\linewidth]{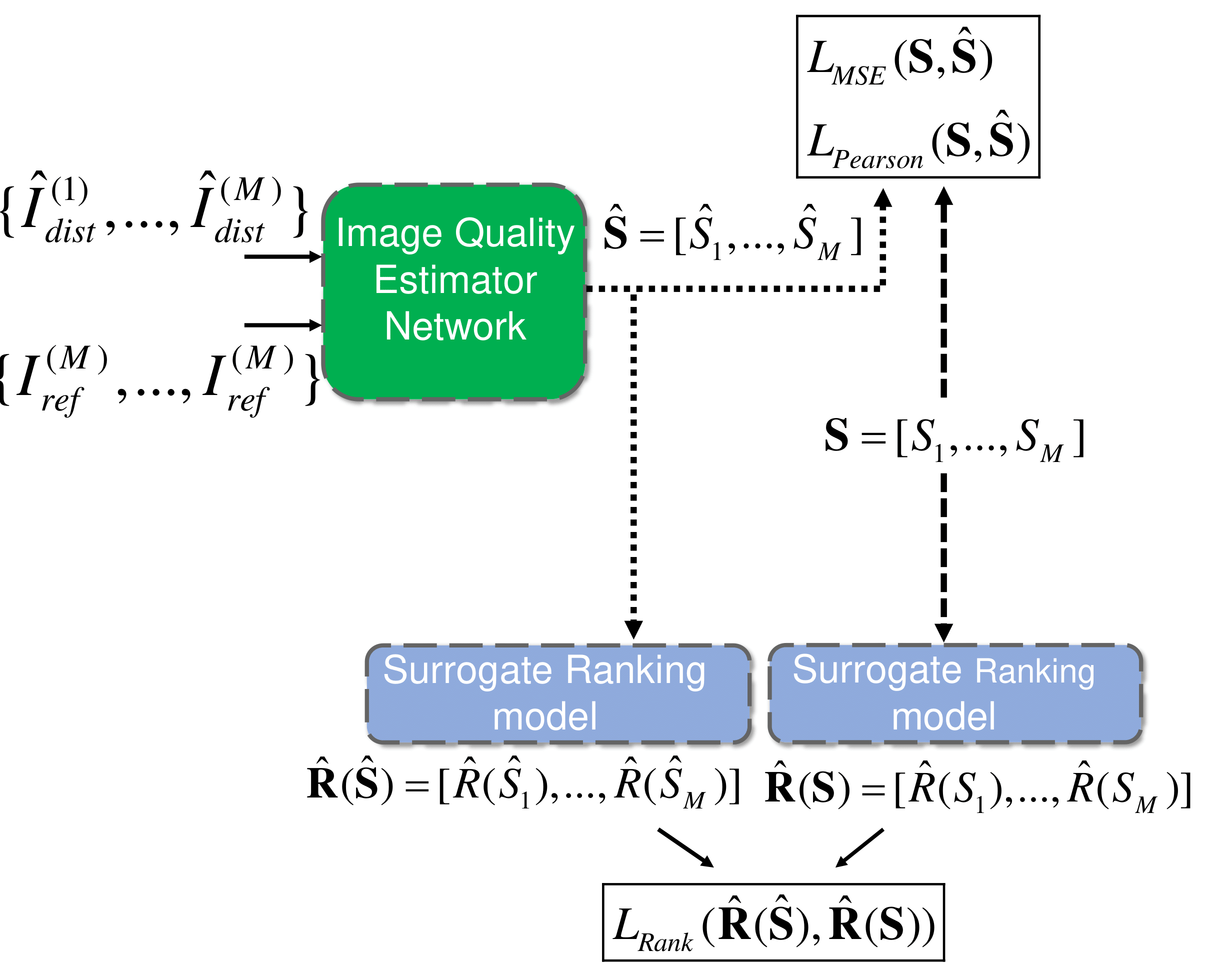}
  \caption{Schematic diagram of using ranking model to compute surrogate ranking loss}
  \label{fig:fig6}
\end{figure}

We have demonstrated the overview of the training procedure in Figure \ref{fig:fig6} to train the IQA network. First we sample $M$ distorted images ($\{\hat{I}_{dist}^{(1)}, \cdots,\hat{I}_{dist}^{(M)}\}$) and their corresponding reference images (($\{I_{ref}^{(1)}, \cdots,I_{ref}^{(M)}\}$). Then we pass these images to the networks to obtain the estimated IQA score ($\Shb$). By having the ground truth MOS scores ($\Sb$), the Pearson loss function and MSE can be computed. Then, we pass $\Sb$ and $\Shb$ to the surrogate ranking network to get the estimated ranking vectors for each. Eventually, the mean squared error between the outputs of the surrogate ranking model for $\Sb$ and $\Shb$ can be used to compute the surrogate ranking loss function. Once all loss functions are calculated, the IQA network can be trained by backpropagation since all operations are differentiable with respect to the IQA network's parameters.  Note that the surrogate ranking model can be fine-tuned during the IQA network training since we can easily compute the true rank vector of $\Shb$ and $\Sb$. 
In the network's evaluation step, we only use IQA network since we are particularly interested in IQA score of the input. Therefore, the surrogate ranking network does not increase the model complexity at the inference time.

\section{Experimental Results}

\subsection{Experiment Setting}
To train the models, we have used PIPAL dataset \cite{pipal}. The dataset contains $1.13$ million images, including the results of GAN-based methods.  Human judgments score the images, and each image's final score is assigned using the "Elo system". We have used an ensemble of models using the ASNA design, the Siamese-Difference model with ConvLSTM, and a Siamese-Difference model without ConvLSTM.
We have used Adam optimizer with learning rate $10^{-4}$ and $\beta_1= 0.9$, $\beta_2= 0.999$ with batch size $16$. The models are first pretrained on TID and PieApp datasets for $20$ epochs. The coefficients for the surrogate loss function is $0.1$, for Pearson correlation coefficient loss and MSE are $0.5$ and $0.5$ respectively. The learning rate is halved each $10$ epochs. We augmented the images by rotating or flipping and scaling the intensity channel in the Lab color space \cite{https://doi.org/10.13140/2.1.1160.2241}. Scaling the intensity channel is in the range of $0.3$ to $1.5$. In the evaluation time, in addition to the model ensembling, we have also used Self Ensambling of the models by rotating, flipping, and scaling intensity channels of the images.

\subsubsection{PLCC vs Distortion Type}
\begin{figure}[h]
  \centering
  \begin{subfigure}[b]{1\linewidth}
    \includegraphics[width=\linewidth]{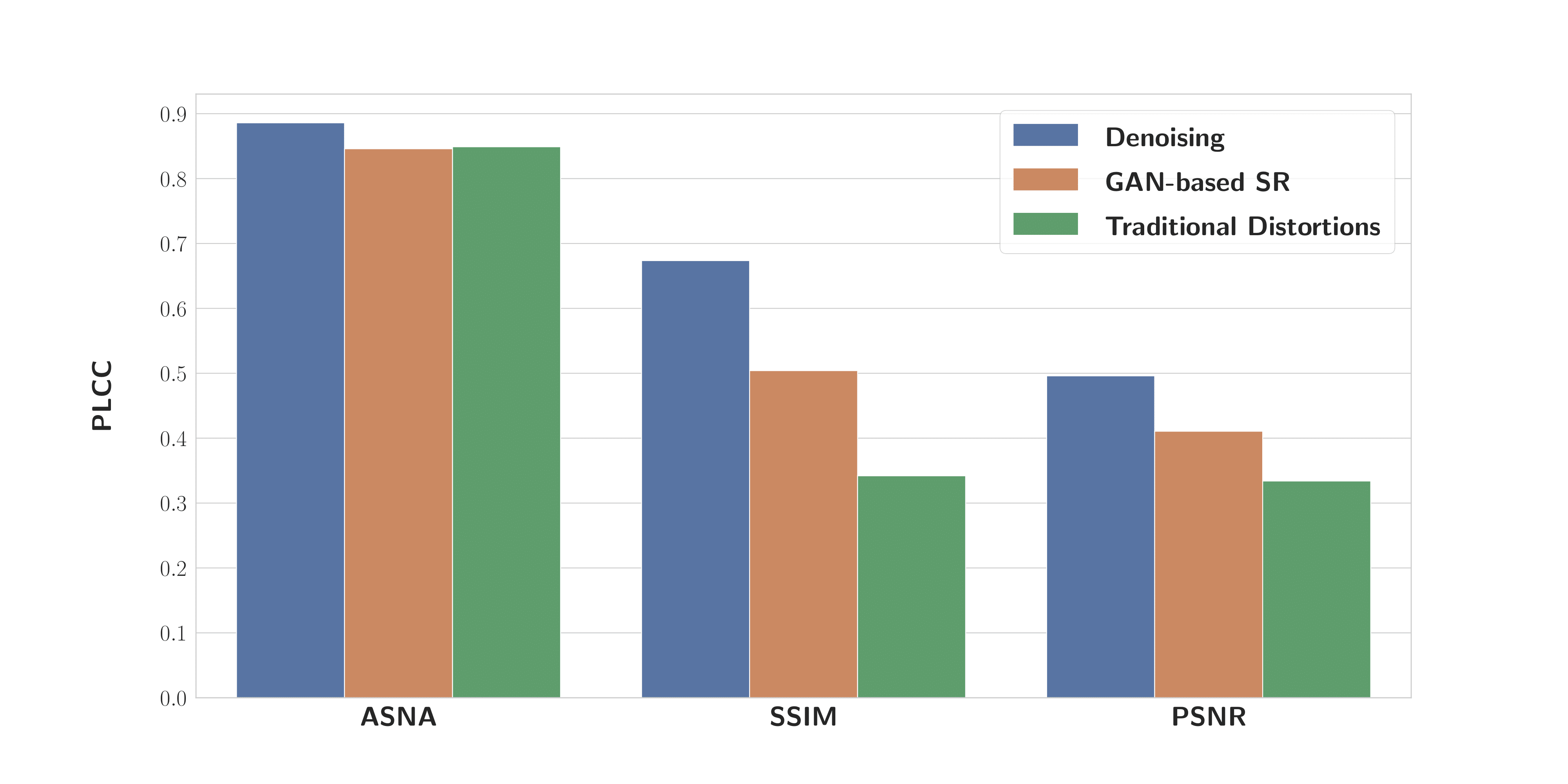}
  \end{subfigure}
  \caption{PLCC for different types of distortions}
  \label{fig:plccdistortion}
\end{figure}

One of the essential features of the proposed metric is that it should perform well on all distortions, especially for the artifacts produced by GANs. We have shown PLCC for different types of distortions in Figure \ref{fig:plccdistortion} for PIPAL dataset. As shown, our method outperformed traditional metrics, including PSNR and SSIM, indicating the capability of ASNA for assessing images that have different distortion characteristics.

\subsection{NTIRE 2021 Perceptual IQA Challenge}
\begin{figure}[h]
  \centering
  \begin{subfigure}[b]{0.49\linewidth}
    \includegraphics[width=\linewidth]{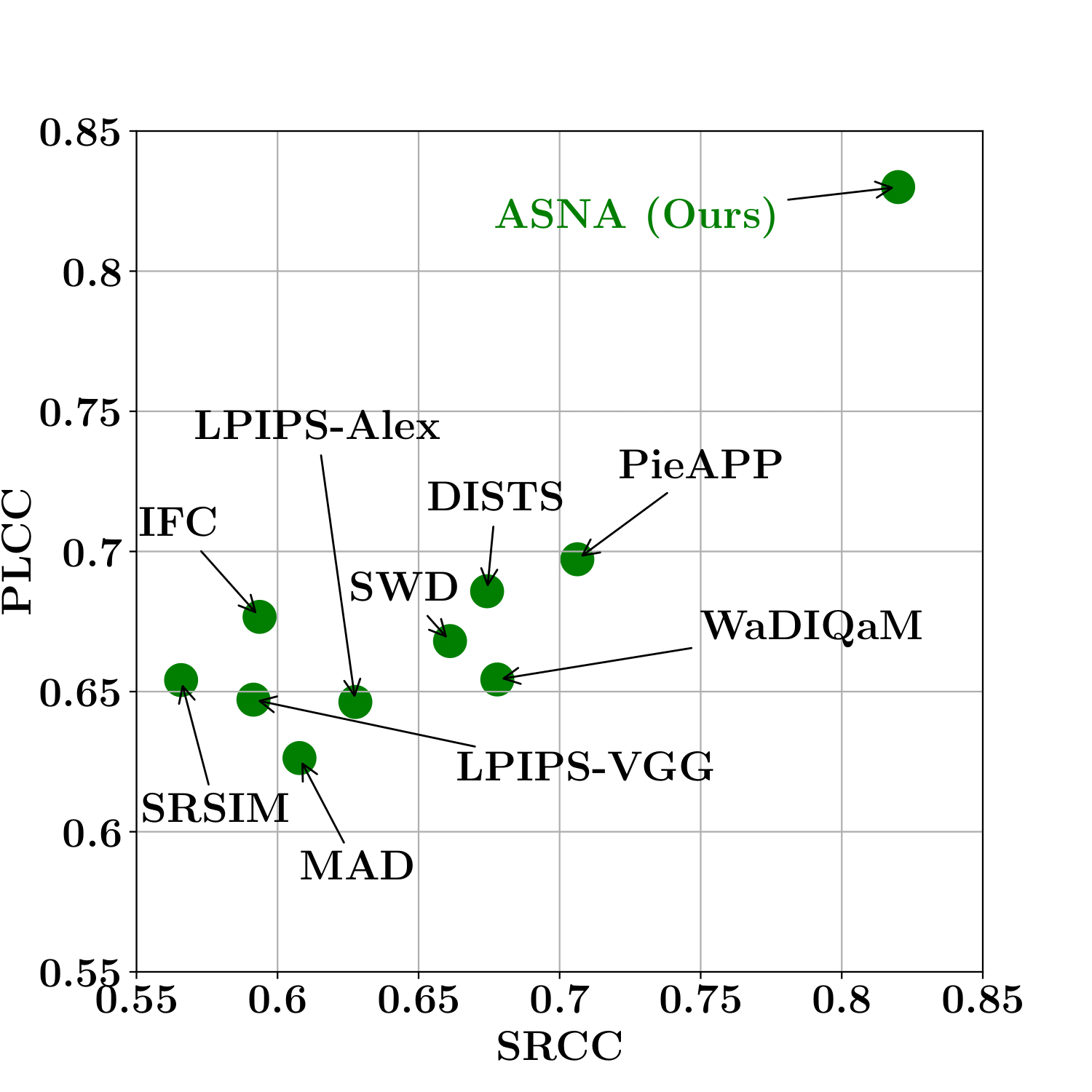}
  \end{subfigure}
  \begin{subfigure}[b]{0.49\linewidth}
    \includegraphics[width=\linewidth]{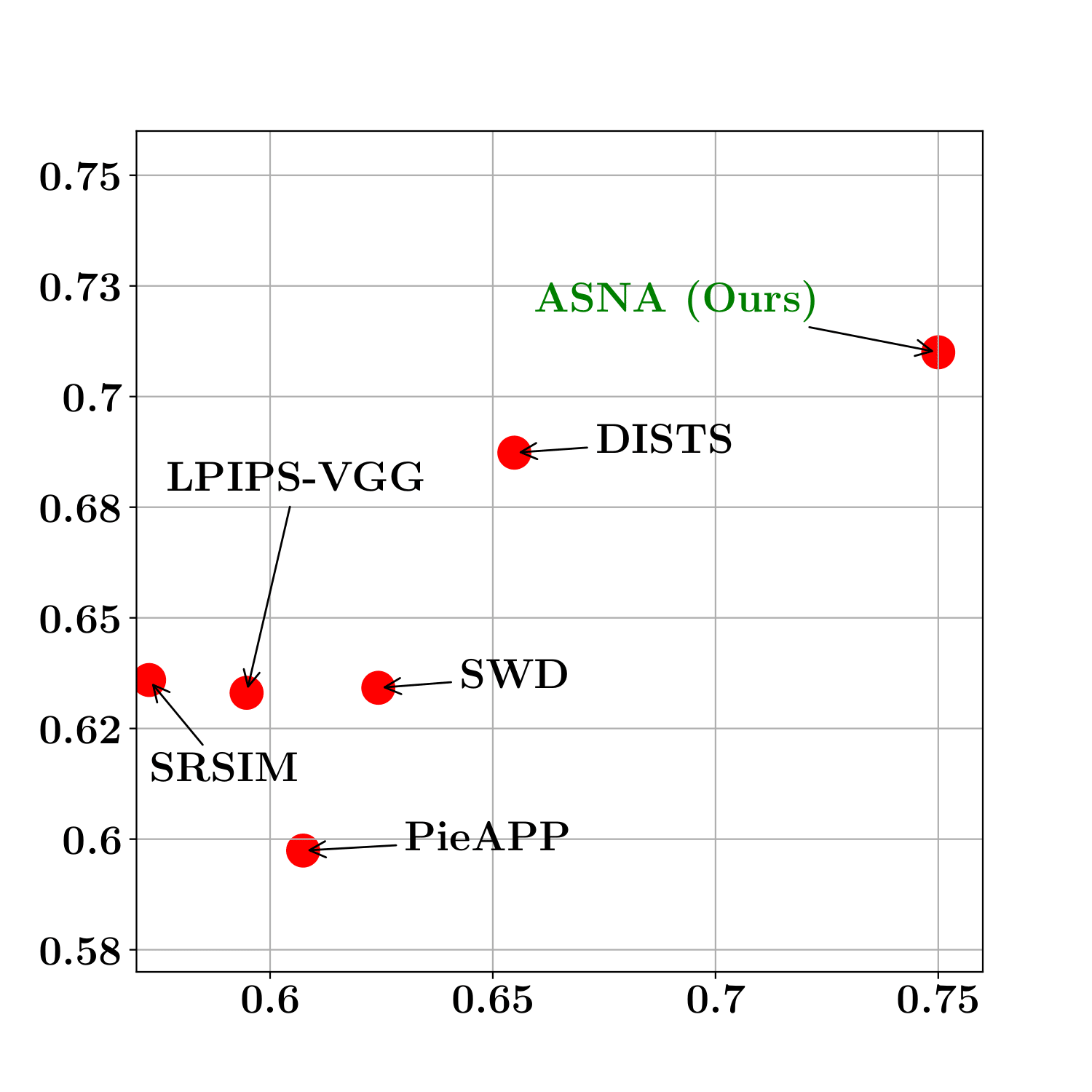}
  \end{subfigure}
  \caption{PLCC vs. SRCC}
  \label{fig:fig8}
\end{figure}

\begin{figure*}[b] 
    \centering
    \includegraphics[width=\linewidth]{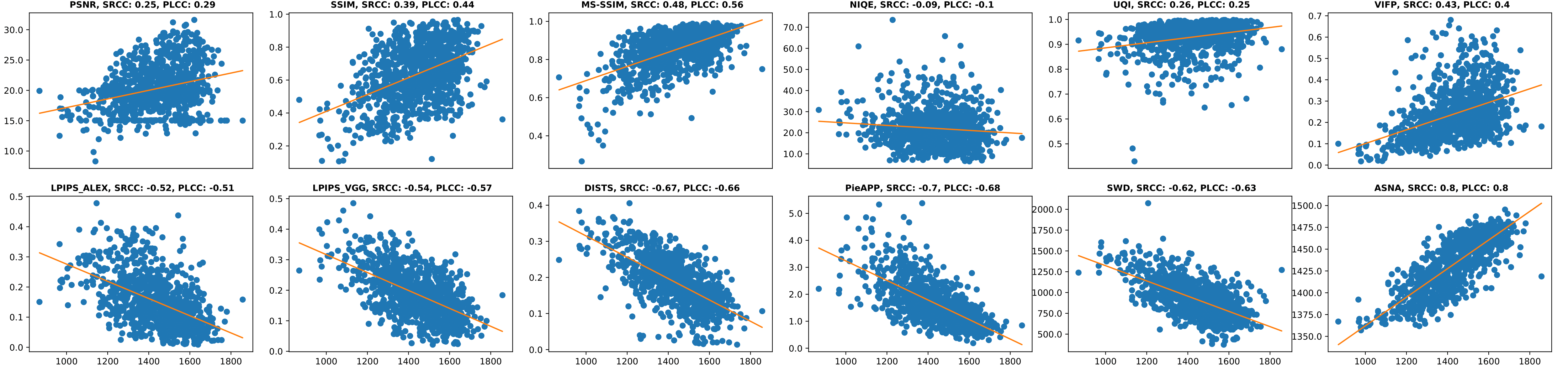}
      \caption{IQA vs. MOS for different methods on PIPAL validation dataset}
	\label{fig:iqamos}
\end{figure*}

We participated in \textbf{\textit{NTIRE 2021 Perceptual IQA}} challenge to evaluate and compare our method to others. The results are demonstrated in Table \ref{table:t1} and Figure \ref{fig:fig8}. As shown, our method achieves superior performance over other approaches to estimate MOS that is well correlated with human judgment on the validation and test dataset of the challenge.
The proposed method outperforms Deep learning-based methods and traditional methods for IQA. To analyze how much the score given by ASNA is aligned with MOS, we have plotted IQA vs. MOS on the PIPAL validation dataset for various methods in Figure \ref{fig:iqamos}. As illustrated, the IQA scores proposed by ASNA are extensively lined up with MOS scores. In addition to that, one can see the variance of the standard measures such as PSNR and SSIM around the fitted line is significantly high, which is an indicator of these measures' poor performance.
\begin{table}[H]
\begin{adjustbox}{width=\columnwidth,center}
\begin{tabular}{  c | c | c | c | c | c | c } 
\hline
 & \multicolumn{3}{c}{\textcolor{Blue}{Validation}} & \multicolumn{3}{c}{\textcolor{OliveGreen}{Test}}  \\
 \hline
\textit{IQA} name & \textcolor{Blue}{\textbf{\textit{Total Score}}} & $|$ \textcolor{Blue}{\textbf{\textit{SRCC}}} $|$ & $|$ \textcolor{Blue}{\textbf{\textit{PLCC}}}$|$ & \textcolor{OliveGreen}{\textbf{\textit{Total Score}}} & $|$ \textcolor{OliveGreen}{\textbf{\textit{SRCC}}} $|$ & $|$ \textcolor{OliveGreen}{\textbf{\textit{PLCC}}} $|$ \\ 
\hline
PSNR\cite{Hore2010} & $0.54$  & $0.26$ & $0.29$ & $0.52$  & $0.24$ & $0.27$ \\ 
\hline
NQM\cite{841940} & $0.76$  & $0.34$ & $0.41$ & $0.75$  & $0.36$ & $0.39$ \\
\hline
UQI\cite{wang2002universal} & $1.03$  & $0.48$ & $0.54$ & $0.86$  & $0.41$ & $0.45$ \\
\hline
SSIM\cite{Wang2004} & $0.73$  & $0.33$ & $0.39$ & $0.75$  & $0.36$ & $0.39$ \\
\hline
MS-SSIM\cite{wang2003multiscale} & $1.04$  & $0.48$ & $0.56$ & $0.96$  & $0.46$ & $0.50$ \\
\hline
IFC\cite{sheikh2005information} & $1.27$  & $0.59$ & $0.67$ & $1.04$  & $0.48$ & $0.55$  \\
\hline
VIF\cite{1576816} & $0.95$  & $0.43$ & $0.52$ & $0.87$  & $0.39$ & $0.47$ \\
\hline
VSNR\cite{4286985} & $0.69$  & $0.32$ & $0.37$ & $0.77$  & $0.36$ & $0.41$ \\
\hline
RFSIM\cite{zhang2010rfsim} & $0.57$  & $0.26$ & $0.30$ & $0.63$  & $0.30$ & $0.32$ \\
\hline
GSM\cite{6081939} & $0.88$  & $0.41$ & $0.46$ & $0.87$  & $0.40$ & $0.46$ \\
\hline
SRSIM\cite{6467149} & $1.21$  & $0.56$ & $0.65$ & $1.20$  & $0.57$ & $0.63$ \\
\hline
FSIM\cite{5705575} & $1.02$  & $0.46$ & $0.56$ & $1.07$  & $0.50$ & $0.57$ \\
\hline
FSIMc\cite{5705575} & $1.02$  & $0.46$ & $0.55$ & $1.07$  & $0.50$ & $0.57$  \\
\hline
VSI\cite{6873260} & $0.96$  & $0.45$ & $0.51$ & $0.97$  & $0.45$ & $0.51$ \\
\hline
MAD\cite{larson2010most} & $1.23$  & $0.60$ & $0.62$ & $1.12$  & $0.54$ & $0.58$ \\
\hline
NIQE\cite{mittal2012making} & $0.16$  & $0.06$ & $0.10$ & $0.16$  & $0.03$ & $0.13$ \\
\hline
MA\cite{ma2017learning} & $0.40$  & $0.20$ & $0.20$ & $0.28$  & $0.14$ & $0.14$ \\
\hline
PI\cite{blau2018perception} & $0.33$  & $0.16$ & $0.16$ & $0.24$  & $0.10$ & $0.14$ \\
\hline
\textcolor{Red}{LIPIS-Alex\cite{zhang2018unreasonable}}
 & $1.27$  & $0.62$ & $0.64$ & $1.13$  & $0.56$ & $0.57$ \\
\hline
\textcolor{Red}{LIPIS-VGG\cite{zhang2018unreasonable}} & $1.23$  & $0.59$ & $0.64$ & $1.22$  & $0.59$ & $0.63$ \\
\hline
\textcolor{Red}{PieApp\cite{prashnani2018pieapp}} & $1.40$  & $0.70$ & $0.69$ & $1.20$  & $0.60$ & $0.59$ \\
\hline
\textcolor{Red}{WaDIQam\cite{bosse2017deep}} & $1.33$  & $0.67$ & $0.65$ & $1.10$  & $0.55$ & $0.54$ \\
\hline 
\textcolor{Red}{DISTS\cite{ding2021comparison}} & $1.36$  & $0.67$ & $0.68$ & $1.34$  & $0.65$ & $0.68$ \\
\hline
\textcolor{Red}{SWD\cite{gu2020image}} & $1.32$  & $0.66$ & $0.66$ & $1.25$  & $0.62$ & $0.63$ \\
\hline
\textcolor{Red}{ASNA (Ours)} & \textcolor{OliveGreen}{$\mathbf{1.65}$}  & \textcolor{OliveGreen}{$\mathbf{0.82}$} & \textcolor{OliveGreen}{$\mathbf{0.83}$} & \textcolor{OliveGreen}{$\mathbf{1.47}$}  & \textcolor{OliveGreen}{$\mathbf{0.75}$} & \textcolor{OliveGreen}{$\mathbf{0.71}$} \\
\hline
\end{tabular}
\end{adjustbox}
\caption{NTIRE 2021 challenge preliminary results (\textcolor{Red}{Red}: Deep Learning-Based methods)}
\label{table:t1}
\end{table}

\subsection{Evaluation on TID 2013 and LIVE Datasets}
To further investigate the generalization of ASNA, we have also evaluated the performance of ASNA on two other benchmark datasets, TID 2013 \cite{PONOMARENKO201557} and LIVE \cite{1709988}. The LIVE database's first release is used to compare ASNA with other IQA methods, which contains different distortions, including compression artifacts and gaussian noise. As can be seen, our method achieved a superior performance on TID 2013 dataset. Note that the number of test images is limited for LIVE dataset; however, our method still has a solid performance on this dataset.

\begin{table}[H]
\begin{adjustbox}{width=\columnwidth,center}
\begin{tabular}{  c | c | c | c | c | c | c } 
\hline
 & \multicolumn{3}{c}{\textcolor{Blue}{TID 2013}} & \multicolumn{3}{c}{\textcolor{OliveGreen}{LIVE}}  \\
 \hline
\textit{IQA} name & \textcolor{Blue}{\textbf{\textit{Total Score}}} & $|$ \textcolor{Blue}{\textbf{\textit{SRCC}}} $|$ & $|$ \textcolor{Blue}{\textbf{\textit{PLCC}}}$|$ & \textcolor{OliveGreen}{\textbf{\textit{Total Score}}} & $|$ \textcolor{OliveGreen}{\textbf{\textit{SRCC}}} $|$ & $|$ \textcolor{OliveGreen}{\textbf{\textit{PLCC}}} $|$ \\ 
\hline
PSNR\cite{Hore2010} & $1.33$  & $0.68$ & $0.65$ & $1.80$  & $0.91$ & $0.90$ \\ 
\hline
SSIM\cite{Wang2004} & $1.37$  & $0.68$ & $0.69$ & $1.89$  & $0.96$ & $0.93$ \\
\hline
MS-SSIM\cite{wang2003multiscale} & $1.54$  & $0.77$ & $0.77$ & $1.85$  & $0.97$ & $0.88$ \\
\hline
UQI\cite{wang2002universal} & $1.76$  & $0.58$ & $0.26$ & $1.61$  & $0.87$ & $0.74$ \\
\hline
VIFP \cite{6411669} & $1.17$  & $0.60$ & $0.57$ & $1.93$  & $0.97$ & $0.95$ \\
\hline
NIQE\cite{mittal2012making} & $0.36$  & $0.19$ & $0.18$ & $0.06$  & $0.01$ & $0.07$ \\
\hline
\textcolor{Red}{LIPIS-Alex\cite{zhang2018unreasonable}}
 & $1.51$  & $0.80$ & $0.71$ & $1.85$  & $0.96$ & $0.90$ \\
\hline
\textcolor{Red}{LIPIS-VGG\cite{zhang2018unreasonable}} & $1.53$  & $0.75$ & $0.78$ & $1.88$  & $0.95$ & $0.93$ \\
\hline
\textcolor{Red}{PieApp\cite{prashnani2018pieapp}} & $1.45$  & $0.83$ & $0.62$ & $1.79$  & $0.93$ & $0.87$ \\
\hline
\textcolor{Red}{DISTS\cite{ding2021comparison}} & $1.48$  & $0.72$ & $0.76$ & \textcolor{OliveGreen}{$\mathbf{1.92}$}  & \textcolor{OliveGreen}{$\mathbf{0.96}$} & \textcolor{OliveGreen}{$\mathbf{0.95}$} \\
\hline
\textcolor{Red}{SWD\cite{gu2020image}} & $1.51$  & $0.75$ & $0.76$ & $1.87$  & $0.95$ & $0.92$ \\
\hline
\textcolor{Red}{ASNA (Ours)} & \textcolor{OliveGreen}{$\mathbf{1.51}$}  & \textcolor{OliveGreen}{$\mathbf{0.73}$} & \textcolor{OliveGreen}{$\mathbf{0.78}$} & $1.84$  & $0.92$ & $0.92$ \\
\hline
\end{tabular}
\end{adjustbox}
\caption{Performance of various methods on LIVE and TID 2013 datasets (\textcolor{Red}{Red}: Deep Learning Based methods)}
\label{table:t2}
\end{table}

\subsection{Visualization of ASNA}
\begin{figure*}[h]
 \centering
  \begin{subfigure}[b]{0.24\linewidth}
    \includegraphics[width=\linewidth]{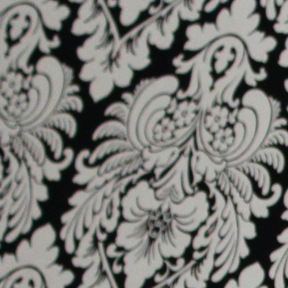}
  \end{subfigure}
    \begin{subfigure}[b]{0.24\linewidth}
    \includegraphics[width=\linewidth]{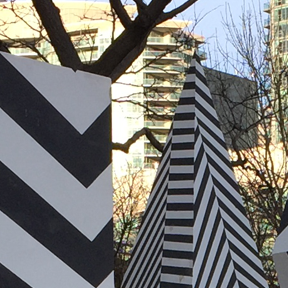}
  \end{subfigure}
    \begin{subfigure}[b]{0.24\linewidth}
    \includegraphics[width=\linewidth]{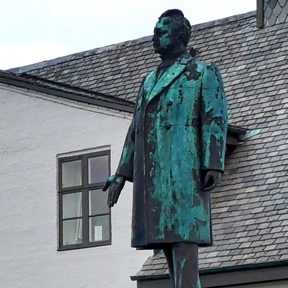}
  \end{subfigure}
    \begin{subfigure}[b]{0.24\linewidth}
    \includegraphics[width=\linewidth]{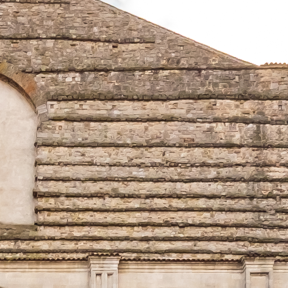}
  \end{subfigure}
    \begin{subfigure}[b]{0.24\linewidth}
    \includegraphics[width=\linewidth]{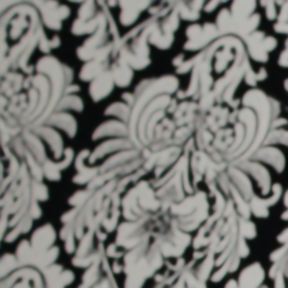}
  \end{subfigure}
    \begin{subfigure}[b]{0.24\linewidth}
    \includegraphics[width=\linewidth]{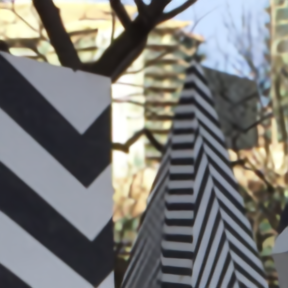}
  \end{subfigure}
    \begin{subfigure}[b]{0.24\linewidth}
    \includegraphics[width=\linewidth]{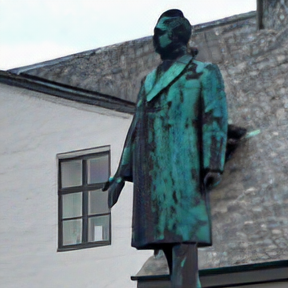}
  \end{subfigure}
    \begin{subfigure}[b]{0.24\linewidth}
    \includegraphics[width=\linewidth]{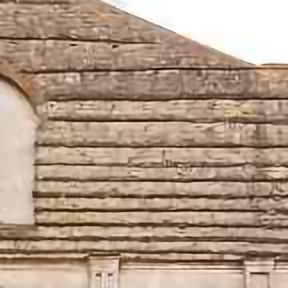}
  \end{subfigure}
    \begin{subfigure}[b]{0.24\linewidth}
    \includegraphics[width=\linewidth]{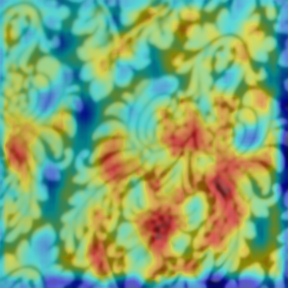}
    \caption{}
  \end{subfigure}
      \begin{subfigure}[b]{0.24\linewidth}
    \includegraphics[width=\linewidth]{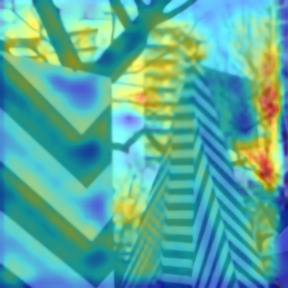}
    \caption{}
  \end{subfigure}
    \begin{subfigure}[b]{0.24\linewidth}
    \includegraphics[width=\linewidth]{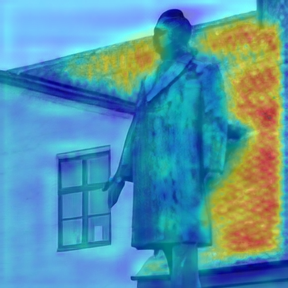}
    \caption{}
  \end{subfigure}
    \begin{subfigure}[b]{0.24\linewidth}
    \includegraphics[width=\linewidth]{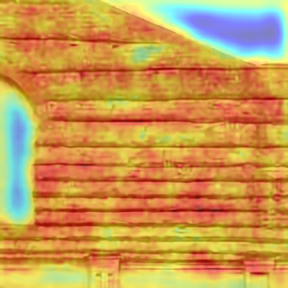}
    \caption{}
  \end{subfigure}
  \caption{\small Visualization of the important parts of the inputs for ASNA for some pairs of images (First row: reference images, second row: distorted images, third row: visualized heat maps using Grad-CAM method)}
  \label{fig:figfin}
\end{figure*}

To understand how ASNA processes the inputs, we have used the last convolutional layer's activations maps. They show the essential features extracted by ASNA for IQA estimation. They have been visualized for some examples using Grad-CAM method \cite{Selvaraju_2019}. The corresponding heat maps and inputs are demonstrated in Figure \ref {fig:figfin}. As illustrated, ASNA focuses on the parts that are visually more important to humans, such as distorted textures and details in the images to estimate IQA scores.

\section{Conclusion}
In this paper, we propose a novel design method for image quality assessment.  The key innovation is to have a suitable design that can catch the subtle difference between the distorted and the reference images. We achieved this goal by an attention-based Siamese-Difference neural network, dubbed ASNA. We have also proposed a surrogate ranking loss function to improve SRCC of the proposed approach. Our proposed full reference IQA is well correlated with subjective human scores for the images.
Experiments show that our method has a significantly greater PLCC and SRCC with MOS compared to other methods for IQA.

\clearpage

{\small
\bibliographystyle{ieee_fullname}
\bibliography{asna}
}

\end{document}